\documentclass[sigconf]{acmart}

\usepackage{microtype}
\usepackage{tikz}

\usetikzlibrary{positioning}
\usepackage{amsmath,amsfonts}

\usepackage{graphicx}
\usepackage{booktabs}
\usepackage{multirow}
\usepackage{tabularx}
\usepackage{adjustbox}

\usepackage{algorithm}
\usepackage{algpseudocode}
\algrenewcommand{\algorithmiccomment}[1]{\hfill$\triangleright$~#1}

\usepackage{listings}
\usepackage{xcolor}
\lstdefinestyle{code}{%
 basicstyle=\ttfamily\small,
 breaklines=true,
 frame=single,
 numbers=left,
 numberstyle=\tiny,
 keywordstyle=\color{blue},
 commentstyle=\color{gray},
 stringstyle=\color{teal}
}
\graphicspath{{./}{figures/}{figs/}{Paper/}}

\newcommand{\safeincludegraphics}[2][]{%
  \IfFileExists{#2}{\includegraphics[#1]{#2}}{%
    \fbox{\parbox[c][1.8in][c]{0.9\columnwidth}{\centering Missing figure file: \texttt{\detokenize{#2}}}}%
  }%
}

\title{VeriTrans: Fine-Tuned LLM-Assisted NL-to-PL Translation via a Deterministic Neuro-Symbolic Pipeline}

\author{Xuan Liu}
\email{xbl5358@psu.edu}
\affiliation{%
  \institution{The Pennsylvania State University}
  \city{University Park}
  \state{Pennsylvania}
  \country{USA}
}

\author{Dheeraj Kodakandla}
\email{djk6439@psu.edu}
\affiliation{%
  \institution{The Pennsylvania State University}
  \city{University Park}
  \state{Pennsylvania}
  \country{USA}
}

\author{Kushagra Srivastva}
\email{kks6306@psu.edu}
\affiliation{%
  \institution{The Pennsylvania State University}
  \city{University Park}
  \state{Pennsylvania}
  \country{USA}
}

\author{Mahfuza Farooque}
\email{mff5187@psu.edu}
\affiliation{%
  \institution{The Pennsylvania State University}
  \city{University Park}
  \state{Pennsylvania}
  \country{USA}
}

\begin{abstract}
\textbf{VeriTrans} is a reliability-first ML system that compiles natural-language requirements into solver-ready logic with validator-gated reliability. The pipeline integrates an instruction-tuned NL$\!\to\!$PL translator, round-trip reconstruction (PL$\!\to\!$NL) used as a high-precision acceptance gate, and canonical PL$\!\to\!$CNF compilation, all executed via fixed API configuration (temperature$=0$; fine-tuning runs use seed$=42$) and per-item artifact logging (prompts, outputs, hashes) to support auditability and replay-driven debugging. On \textbf{SatBench} (2{,}100 specifications), VeriTrans achieves 94.46\% SAT/UNSAT correctness and 87.73\% median round-trip similarity. Compact fine-tuning on 100--150 curated examples improves fidelity by about 1--1.5\,pp without increasing latency (mean 25.8\,s/spec on our 201-spec runtime subset). A thresholded acceptance policy on the round-trip score exposes a reliability--coverage knob: at $\tau{=}75$, roughly 68\% of items are retained with $\sim$94\% correctness on the accepted set. Validator overhead contributes $<15\%$ of end-to-end runtime, and all prompts/responses and timing metadata are logged to enable replay-driven debugging and regression testing. By separating learned translation from symbolic verification and enforcing deterministic, validator-gated acceptance, VeriTrans turns NL$\!\to\!$logic front-ends into auditable, reproducible components for reliability-critical workflows.
\end{abstract}


\begin{document}
\maketitle


\section{Introduction}
\label{Introduction}

Formal verification pipelines underpin modern hardware, software, and autonomous systems, where correctness failures can propagate across safety-critical components~\cite{DUAN2025115368}. These pipelines rely on SAT-based solvers to validate logical specifications, yet the translation from natural-language (NL) requirements to solver-ready logic remains a major throughput bottleneck. Manual encoding or rule-based parsers often introduce semantic drift and prevent large-scale automation across evolving codebases. 

\textbf{VeriTrans} addresses this gap by automating the full translation path---from natural language (NL) to propositional logic (PL) and deterministically to CNF---while enforcing solver-safety, auditability, and conservative acceptance via validator signals and deterministic back-end compilation/solving. This bottleneck dominates end-to-end verification latency in industrial settings, limiting solver utilization and continuous integration throughput.

Large Language Models (LLMs) have shown significant promise in tasks requiring translation, reasoning, and code generation, offering a potential pathway to automate the NL-to-logic translation bottleneck~\cite{ryu2025divide}. However, their application in the rigorous context of formal verification remains underdeveloped, primarily due to concerns about reliability and the potential for subtle errors or ``hallucinations'' in safety-critical scenarios~\cite{huang2025survey}. 

We illustrate this challenge through a simple specification example that highlights the semantic gap between human-readable requirements and solver-ready representations. Consider the requirement: 
\textit{``If the temperature sensor detects a reading above the threshold and the cooling system is offline, then an emergency shutdown must be triggered or an alarm must sound.''} 
Correctly translating this requires identifying atomic propositions (e.g., \texttt{T} for temperature high, \texttt{C} for cooling offline, \texttt{S} for shutdown, \texttt{A} for alarm) and capturing the logical structure: \((T \land C) \to (S \lor A)\). Traditional methods involving manual translation or rigid templates often fail to handle the nuances and variability inherent in natural language. Automating this translation promises significant efficiency gains; a system achieving high correctness (e.g., $>90\%$), even if requiring human review for ambiguous cases, could drastically reduce development bottlenecks while maintaining high reliability.

To bridge this gap, we introduce \textbf{VeriTrans}, an end-to-end pipeline designed to automate the translation of NL specifications into CNF for reliable SAT-based verification. VeriTrans integrates fine-tuned LLMs for the initial NL$\to$PL step with deterministic symbolic methods for PL$\to$CNF conversion. A core feature of our system is a round-trip consistency validator (PL$\to$NL) that provides a conservative consistency signal used as a high-precision acceptance gate under a fixed vocabulary and prompt schema. Evaluating VeriTrans on the comprehensive SatBench dataset, our system achieves 94.46\% overall SAT/UNSAT correctness (99.81\% accuracy on SAT instances, 89.13\% on UNSAT instances) with a median round-trip semantic similarity of 87.73\%. This work demonstrates that a synergistic combination of LLM fine-tuning and careful systems engineering can effectively address a long-standing challenge in scalable formal verification.

We emphasize that this work does not attempt to solve the problem of semantic correctness in natural language--to--logic translation. Natural language specifications are inherently ambiguous, and multiple logically distinct formulas may plausibly correspond to the same description. Determining semantic equivalence between natural language and formal logic generally requires domain knowledge or human judgment and is outside the scope of this paper. Instead, we study how LLM-generated logical formulas can be safely integrated into verification pipelines by prioritizing reproducibility, auditability, and conservative failure modes.

\subsection{Motivation}

The translation from informal natural-language specifications to precise formal logic remains a primary obstacle in the widespread adoption of formal verification. Manual translation is inherently slow and susceptible to subtle human errors that can undermine the entire verification process. While LLMs present a scalable alternative, their general-purpose nature necessitates domain adaptation to achieve the required precision for formal methods.

\subsection{Challenges}

Applying LLMs to the demanding domain of formal verification introduces several critical challenges:

\begin{itemize}
 \item \textbf{Logical Equivalence and Scope Control:} The generated logical formula should faithfully capture the NL specification within a constrained propositional vocabulary and consistent operator scope. Subtle inaccuracies---such as misplaced negations or incorrect operator precedence---can lead to erroneous verification outcomes, potentially accepting unsafe systems or rejecting correct ones.
 \item \textbf{Automatic Validation and Failure Prediction:} LLM outputs can be fluent yet mis-scoped or inconsistent with the intended constraint structure, motivating conservative validators that identify low-confidence translations for rejection or review. In formal verification, a hallucinated formula might be syntactically valid but logically flawed, posing significant risks in safety-critical applications~\cite{wei2022cot}.
 \item \textbf{Dataset and Solver Throughput:} Real-world system specifications can be complex, involving nested clauses, intricate conditional dependencies, and specialized terminology. The translation system must robustly handle this variability while maintaining high accuracy.
 \item \textbf{Confidence-Aware Acceptance Policies:} For safety-critical use, high average performance is insufficient. The system needs mechanisms to identify potentially low-confidence or ambiguous translations, flagging them for necessary human oversight to ensure overall dependability.
\end{itemize}

\subsection{Contributions}

This paper presents \textbf{VeriTrans}, a fine-tuned NL$\to$PL$\to$CNF$\to$SAT pipeline optimized for accuracy and efficiency in formal verification workflows. Our key contributions include:

\begin{itemize}
 \item \textbf{A Systematic Hybrid Pipeline:} We designed and implemented an end-to-end system combining LLM-based NL-to-PL translation with deterministic symbolic conversion (PL-to-CNF) and SAT solving, crucially integrating a round-trip (PL-to-NL) consistency check for validation.
 \item \textbf{Effective Fine-Tuning for Formal Logic:} We demonstrate that targeted, small-scale supervised fine-tuning (using as few as 100--150 examples) significantly enhances the LLM's ability to translate NL specifications into accurate PL formulas. On the full 2{,}100-item SatBench dataset, VeriTrans achieves 94.46\% overall SAT/UNSAT correctness and a median round-trip similarity of 87.73\%.
 \item \textbf{Quantitative Hallucination Detection:} We propose and validate a practical mechanism for detecting potential LLM hallucinations in this context by translating the generated PL back to NL and measuring semantic similarity against the original input, providing a quantifiable measure of translation fidelity.
 \item \textbf{Comprehensive Evaluation on SatBench:} We perform extensive experiments using the SatBench benchmark~\cite{wei2025satbench}, analyzing performance across various complexity levels, identifying common failure modes, and demonstrating efficiency benefits (throughput and latency) compared to purely manual approaches.
\end{itemize}

Together, these results demonstrate how reliability-first ML design can transform formal verification pipelines into reproducible, high-throughput systems. The remainder of this paper details the related work and background (Section~\ref{sec:relatedwork}, Section~\ref{sec:background}), presents our methodology (Section~\ref{sec:methodology}), discusses the results (Section~\ref{sec:results}), and concludes with future directions (Section~\ref{sec:conclusion}).

\section{Background}
\label{sec:background}

SAT-based formal verification validates whether logical constraints derived from system specifications can be satisfied simultaneously. 
A specification is typically represented in \textit{Conjunctive Normal Form} (CNF), where formulas are conjunctions of disjunctions of literals. 
A modern SAT solver determines satisfiability by assigning truth values to variables such that all clauses evaluate to true. 
While solver performance has improved dramatically, the upstream step -- translating human-readable specifications into machine-verifiable logic -- remains a major bottleneck.

In VeriTrans, we formalize this translation process as a four-stage pipeline: 
(1) natural language (NL) $\rightarrow$ propositional logic (PL), 
(2) PL $\rightarrow$ natural language (PL$\rightarrow$NL) for round-trip validation, 
(3) PL $\rightarrow$ CNF for solver compatibility, and 
(4) CNF $\rightarrow$ SAT outcome determination. 
This mapping allows the system to detect and flag inconsistencies at the language--logic interface before solver execution.

Determinism here refers to the symbolic back-end (PL$\to$CNF$\to$SAT) and artifact logging for the learned stages. We execute LLM calls with a fixed decoding configuration (temperature$=0$) and log prompts, outputs, and timing to support replay-driven debugging and regression testing alongside fully deterministic CNF compilation and solver traces.

The validator-gated acceptance policy employs a threshold~$\tau$ on round-trip similarity to balance fidelity and coverage. 
Lower thresholds allow more coverage at the expense of potential drift, while higher thresholds favor correctness and auditability. 
This design ensures that the learned front-end interacts predictably with the symbolic back-end.

Building on these foundations, the next section presents the complete VeriTrans methodology, describing its architecture, fine-tuning strategy, and deterministic validation loop.

\section{Methodology}
\label{sec:methodology}

All LLM calls in VeriTrans are executed with a fixed decoding configuration (temperature~=~0) and comprehensive artifact logging to support replay-driven debugging, while the PL$\to$CNF$\to$SAT stages are fully deterministic.
This section outlines the design principles, pipeline stages, fine-tuning setup, and evaluation metrics of our system.

\subsection{Design Goals and Problem Framing}
\label{sec:design-goals}
We aim to construct a lightweight, reproducible ML system that translates informal specifications into formal logic while minimizing semantic drift and hallucination. 
Concretely, the pipeline must (i) map natural language (NL) into a scoped propositional logic (PL) vocabulary, 
(ii) deterministically convert PL to conjunctive normal form (CNF) suitable for SAT solvers, and 
(iii) validate semantic fidelity using round-trip consistency and structural sanity checks. 
These goals emphasize \emph{correctness under resource constraints}; therefore, we evaluate both logical accuracy and system metrics such as latency, token cost, and throughput.

\subsection{System Overview}
\label{sec:system-overview}
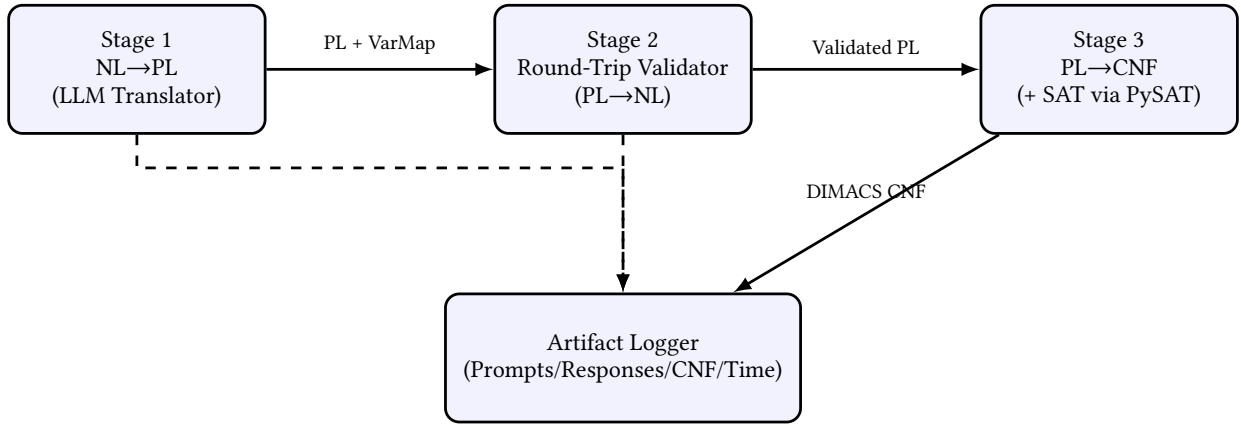
\begin{figure*}[t]
\centering
\resizebox{0.92\textwidth}{!}{%
\begin{tikzpicture}[node distance=2.3cm,>=latex,auto,thick]
\tikzstyle{block}=[rectangle,draw,fill=blue!5,rounded corners,
minimum height=1.3cm,minimum width=2.6cm,align=center,font=\footnotesize]
\tikzstyle{line}=[draw,->,thick]

\node[block] (nl2pl) {Stage~1\\NL$\rightarrow$PL\\(LLM Translator)};
\node[block,right=of nl2pl] (validator) {Stage~2\\Round-Trip Validator\\(PL$\rightarrow$NL)};
\node[block,right=of validator] (cnf) {Stage~3\\PL$\rightarrow$CNF\\(+ SAT via PySAT)};
\node[block,below=1.6cm of validator,minimum width=3.6cm] (log) {Artifact Logger\\(Prompts/Responses/CNF/Time)};

\draw[line] (nl2pl) -- node[above,yshift=1pt,font=\scriptsize]{PL + VarMap} (validator);
\draw[line] (validator) -- node[above,yshift=1pt,font=\scriptsize]{Validated PL} (cnf);
\draw[line] (cnf) -- node[above,yshift=1pt,font=\scriptsize]{DIMACS CNF} (log);
\draw[line,dashed] (validator) -- ++(0,-1.0) -| (log);
\draw[line,dashed] (nl2pl) -- ++(0,-1.0) -| (log);
\end{tikzpicture}}
\vspace{-1mm}
\small
\caption{Hybrid neural--symbolic VeriTrans pipeline. 
The LLM performs NL$\!\rightarrow$PL translation; validators enforce linguistic and structural consistency; deterministic PL$\!\rightarrow$CNF conversion and SAT solving provide reproducible verification.}
\label{fig:veritrans_pipeline}
\vspace{-3mm}
\end{figure*}

Figure~\ref{fig:veritrans_pipeline} shows the complete VeriTrans workflow. 
Users provide an English requirement, an optional variable mapping, and contextual information. 
Stage~1 performs NL$\!\rightarrow\!$PL translation using an instruction-tuned LLM. 
Stage~2 reconstructs PL$\!\rightarrow\!$NL text to enable round-trip consistency evaluation. 
Stage~3 converts PL to CNF deterministically via a Tseitin encoder and executes SAT solving. 
Stage~4 applies validator passes -- textual similarity, symbol coverage, and clause-level sanity checks -- under a tunable $\tau$-threshold acceptance policy.

\subsection{Dataset and Preprocessing}
We evaluate VeriTrans on \textbf{SatBench}, a corpus of English specifications paired with gold PL, CNF, and SAT/UNSAT labels. 
Each instance includes: (i) requirement text, (ii) scenario stub, (iii) symbol mapping for entities and predicates, and (iv) ground-truth PL and CNF. 
Preprocessing resolves co-reference, expands shorthand (e.g., ``iff''), and removes annotator artifacts. 
A per-item symbol inventory~$\mathcal{V}$ constrains the NL$\!\rightarrow\!$PL generation space.

\subsection{Stage~1: NL$\rightarrow$PL Translation}
We prompt \texttt{GPT-4o-mini} with a structured schema that lists the allowed vocabulary~$\mathcal{V}$ and requests a single well-formed propositional logic formula~$\varphi$ with a variable mapping~$M$. 
Prompts enforce: (i) use only declared symbols, (ii) explicit negation, and (iii) fully parenthesized operator precedence. 
Inference uses a fixed decoding configuration (temperature~=~0) with one output per input under a fixed prompt schema.
Outputs are parsed and normalized; unparseable items are flagged as rejected and counted toward coverage, with accepted-set metrics reported separately under the $\tau$-gated policy.
When fine-tuned variants are used, they rely on $(x,\varphi)$ pairs drawn from SatBench subsets; this stage performs inference only. 
Section~\ref{sec:results} reports fidelity, latency, and token-efficiency trade-offs.

\subsection{Stage~2: PL$\rightarrow$NL Translation}
Given a formula~$\varphi$ and its mapping~$M$, the model reconstructs a natural-language description via a deterministic prompt. 
Variable aliases (e.g., \texttt{x\_2\_0}$\!\mapsto$``speed at city'') are prepended to the prompt to ensure consistent verbalization. 
The LLM restates $\varphi$ while preserving logical connectives and negation scope. 
Outputs are post-processed for casing and punctuation normalization. 
We compute TF--IDF cosine similarity between the reconstructed text and original requirement~$x$, yielding the round-trip similarity metric used by validators in Section~\ref{sec:stage4-validators}. 
All PL$\!\rightarrow$NL calls are logged to enable replay-driven debugging and regression testing.

\subsection{Stage~3: PL$\rightarrow$CNF Conversion}
Each propositional formula~$\varphi$ is deterministically compiled into an equisatisfiable CNF formula~$\psi$ using a Tseitin encoder. 
Connectives ($\Rightarrow$, $\Leftrightarrow$) are normalized to $\{\lnot,\land,\lor\}$, and auxiliary variables are introduced for sub-formulas. 
CNF is emitted in DIMACS format (\texttt{p cnf \#vars \#clauses}) and solved via PySAT to obtain SAT/UNSAT labels. 
This stage is fully symbolic and deterministic across runs.

\subsection{Stage~4: Validators and Acceptance Policy}
\label{sec:stage4-validators}
We deploy two validator families: 
(1) \textbf{Round-trip:} translate $\varphi$ back to NL, compute TF--IDF similarity with the original input~$x$, and record the TF--IDF cosine similarity as a percentage. 
(2) \textbf{Structural:} check CNF well-formedness, symbol coverage, and clause-level anti-patterns (e.g., tautologies). 
The $\tau$-threshold acceptance policy rejects outputs failing either validator, trading coverage for higher fidelity.

\subsection{Fine-Tuning Approach}
We specialize \texttt{GPT-4o-mini} for NL$\!\rightarrow\!$PL translation using OpenAI's managed supervised fine-tuning (SFT) API , with the training seed fixed at 42 for all fine-tuning jobs. 
Four subsets (\texttt{ft50}, \texttt{ft100}, \texttt{ft150}, \texttt{ft300}) were derived from SatBench to control domain variation. 
Each subset trained a forward (NL$\!\rightarrow\!$PL) and reverse (PL$\!\rightarrow\!$NL) model to enforce bidirectional consistency and scalable evaluation.

\paragraph{Dataset Preparation.}
Each example is formatted in JSON Lines (\texttt{.jsonl}) following OpenAI's ChatML convention~\cite{openai_finetuning}. 
Every record contains a list of \texttt{messages} with roles \texttt{system}, \texttt{user}, and \texttt{assistant}. 
The \texttt{system} message defines the task and output format; 
the \texttt{user} message includes the natural-language scenario and variable mapping; 
the \texttt{assistant} message provides the correct propositional logic formula. 
This conversational structure explicitly teaches the model the transformation objective.

\lstset{
 basicstyle=\ttfamily\small,
 breaklines=true,
 frame=single,
 postbreak=\mbox{\textcolor{red}{$\hookrightarrow$}\space},
 showstringspaces=false,
 captionpos=b,
}

\begin{lstlisting}[float, caption={Example training instance in ChatML format for fine-tuning. The model learns to map the user's prompt to the corresponding propositional formula.}, label={fig:finetune-example}]
<|im_start|>user
Convert the following scenario into a formal logic formula:

If the temperature sensor detects a reading above the threshold
and the cooling system is offline, then an emergency shutdown must
be triggered or an alarm must sound.
<|im_end|>
<|im_start|>assistant
(~x(0,0) V ~x(0,1) V x(0,2) V x(0,3))
<|im_end|>
\end{lstlisting}

\paragraph{Fine-Tuning Process via API.}
The workflow proceeds through four automated steps:
\begin{enumerate}
 \item \textbf{Data Upload:} the curated \texttt{.jsonl} file is uploaded via the OpenAI Files API. 
 \item \textbf{Job Creation:} a fine-tuning job is initiated specifying the base model (\texttt{GPT-4o-mini}) and dataset ID; OpenAI automatically handles validation and hyperparameter selection~\cite{openai_finetuning}. 
 \item \textbf{Model Deployment:} upon completion, a unique identifier for the fine-tuned model is returned. 
 \item \textbf{Inference:} Inference uses the Chat Completions API with temperature~=~0; all prompts/responses are logged for replay and regression testing.
\end{enumerate}

This managed API approach enables domain specialization without maintaining training infrastructure.

\subsection{Evaluation Metrics}
We report four metrics: 
(i) \textbf{SAT/UNSAT correctness}, the match between solver output and gold labels; 
(ii) \textbf{Round-trip similarity}, the median TF--IDF cosine similarity between original~$x$ and reconstructed~$y$; 
(iii) \textbf{Latency and token cost}, measured end-to-end; and 
(iv) \textbf{Throughput}, items processed per minute under provider rate limits.

\subsection{Algorithmic Pipeline}
\label{app:algorithms}
Algorithms~\ref{alg:nl2pl}--\ref{alg:pl2cnf} outline the three deterministic stages of
the NL$\!\rightarrow\!$PL$\!\rightarrow\!$CNF pipeline used by \textbf{VeriTrans}.
Each stage produces structured CSV outputs, ensuring reproducibility.
\vspace{4pt}
\begingroup
\setlength{\textfloatsep}{8pt plus 2pt minus 2pt}
\setlength{\floatsep}{6pt plus 2pt minus 2pt}
\setlength{\intextsep}{6pt plus 2pt minus 2pt}

\begin{algorithm}[t]
\caption{Stage~1: NL$\rightarrow$PL (Deterministic Prompting and Extraction)}
\label{alg:nl2pl}
\begin{algorithmic}[1]
\Require Rows with \texttt{conditions}, optional \texttt{variable\_mapping}, and \texttt{scenario}
\Ensure CSV rows with \texttt{generated\_formula}, \texttt{generated\_mapping}, timing, and tokens
\ForAll{row}
 \State $x \gets$ join/parse(\texttt{conditions})
 \State $M_{\mathrm{seed}} \gets$ \texttt{variable\_mapping},~~$S \gets$ \texttt{scenario}
 \State $\textbf{Prompt} \gets \texttt{NL2PL\_PROMPT}\{S, M_{\mathrm{seed}}, x\}$
 \State $(o, t, id, p, c, z) \gets \textsc{CallOpenAI}(\text{Prompt};~T{=}0)$
 \State $(\hat{M}, \hat{\varphi}) \gets \textsc{ExtractMappingAndFormula}(o)$
 \State Write row $\{\hat{\varphi}, \hat{M}, t, p, c, z,\ldots\}$ to CSV
\EndFor
\end{algorithmic}
\end{algorithm}


\begin{algorithm}[t]
\caption{Stage~2: PL$\rightarrow$NL Reconstruction}
\label{alg:pl2nl}
\begin{algorithmic}[1]
\Require Rows $(x, \varphi, M)$
\Ensure Reconstructed text $\hat{y}$ and TF--IDF cosine $s$
\ForAll{rows}
 \State $\varphi \gets$ \texttt{generated\_formula} or skip if empty
 \State Build $M$ text by merging \texttt{input\_mapping} and parsed \texttt{generated\_mapping}
 \State Fill \texttt{PL2NL\_PROMPT} with $\{M, \varphi\}$ and call LLM with $T{=}0$
 \State $\hat{y} \gets$ lines under ``\texttt{Reconstructed Conditions:}''
 \State $s \gets 100 \times \cos(\text{tfidf}(x), \text{tfidf}(\hat{y}))$
 \State Record $(\hat{y}, s)$ and timing/token totals
\EndFor
\State \Return $\{(\hat{y}_i, s_i)\}$
\end{algorithmic}
\end{algorithm}


\begin{algorithm}[t]
\caption{Stage~3: PL$\rightarrow$CNF$\rightarrow$SAT}
\label{alg:pl2cnf}
\begin{algorithmic}[1]
\Require Rows with formula string $\varphi$
\Ensure SAT label (\texttt{SAT}/\texttt{UNSAT}) and DIMACS CNF per row
\ForAll{row}
 \State $\varphi \gets$ row[\texttt{generated\_formula}] or skip if empty
 \State $\varphi \gets \textsc{CanonicalizeIndexedVars}(\varphi)$ \Comment{$x(i,j)\!\to\!x\_i\_j$}
 \State $\varphi \gets \textsc{NormalizeSymbols}(\varphi)$ \Comment{$\lnot,\land,\lor,\to,\leftrightarrow$ $\to$ \texttt{!,\&,|,->,<->}}
 \State $toks \gets \textsc{Tokenize}(\varphi)$;~~$rpn \gets \textsc{ToRPN}(toks)$
 \State $ast \gets \textsc{RPNtoAST}(rpn)$
 \State $(C_{\mathrm{str}}, top) \gets \textsc{TseitinCNF}(ast)$
 \State $(C_{\mathrm{int}}, sym2id) \gets \textsc{MapLitsToInts}(C_{\mathrm{str}})$
 \State $\text{sat} \gets \textsc{MiniSAT22}(C_{\mathrm{int}})$
 \State $\text{dimacs} \gets \textsc{ToDIMACS}(C_{\mathrm{int}}, sym2id)$
 \State Write $\{\texttt{pred\_from\_script}, \texttt{cnf\_dimacs}\}$ to CSV
\EndFor
\end{algorithmic}
\end{algorithm}
\endgroup


\subsection{Round-Trip Consistency Analysis}
\label{sec:roundtrip}
Round-trip validation measures whether a generated logical formula is self-consistent with respect to the system's translation and reconstruction processes. Given a natural language input N, the system generates a propositional logic formula F, reconstructs a natural language description $\hat{N}$ from F using a fixed prompt, and compares N and $\hat{N}$ using a similarity metric. This procedure does not test semantic equivalence between N and F; rather, it evaluates whether the translation remains stable under a constrained, deterministic translation--reconstruction loop.\\
Round-trip similarity quantifies how closely the reconstructed NL matches the original specification after the NL$\!\rightarrow\!$PL$\!\rightarrow\!$NL cycle. 
Scores lie in $[0,100]$ as TF--IDF cosine similarity percentages. 
High similarity is treated as a conservative consistency signal under the fixed prompt and vocabulary constraints.

\begin{table}[t]
 \centering
 \caption{High-consistency mass at $\tau=75\%$.}
 \label{tab:rt-sim-mass}
 \setlength{\tabcolsep}{4pt}
 \small
 \begin{tabular}{lcc}
 \toprule
 Threshold & Count & Proportion \\
 \midrule
 $\ge75\%$ & 1422 & 0.677 \\
 \bottomrule
 \end{tabular}
 \vspace{-4pt}
\end{table}

Across 2{,}100 specifications, about two-thirds exceed 75\% similarity, 
and the median (87.73\%) shows that half deviate by less than 12.27\%. 
The mean of 80.20\% lies within a 95\% confidence interval $[79.24,\,81.16]$, 
and Hoeffding's inequality~\cite{hoeffding1963probability} bounds the probability of deviation by more than 5\% to below $6\times10^{-5}$. 
These results confirm that semantic drift -- and thus hallucination -- in the NL$\!\rightarrow$PL step remains low, 
and that the validator loop provides a reproducible safeguard for correctness under bounded stochastic variation.\\
A low round-trip similarity score does not imply that the generated formula is logically incorrect. A formula may correctly encode one plausible interpretation of the input while still failing reconstruction due to ambiguity, underspecification, or information loss. Such cases constitute intentional false negatives. VeriTrans is designed to favor high precision over recall, prioritizing rejection of uncertain translations rather than speculative acceptance.\\
The round-trip consistency signal is inherently system-relative. Its behavior depends on a fixed logical vocabulary, deterministic decoding parameters, and a predefined reconstruction prompt. Consequently, the signal should not be interpreted as model-agnostic or semantics-aware. Its purpose is to provide a conservative, reproducible validation mechanism within a controlled pipeline, not a general test of logical equivalence.


\section{Results}
\label{sec:results}

This section evaluates \textbf{VeriTrans} across three dimensions:
(\textit{i}) translation fidelity and correctness,
(\textit{ii}) runtime and throughput efficiency,
and (\textit{iii}) reliability under validator-gated acceptance. Unless otherwise stated, results are reported on the full SatBench dataset (2,100 specifications); Table~\ref{tab:ft-summary} summarizes experiments on a 201-spec subset used for controlled fine-tuning comparisons.

\vspace{3pt}
\subsection{Fine-Tuning Efficiency--Fidelity Trade-off}

Table~\ref{tab:ft-summary} and Fig.~\ref{fig:ft_scaling_curve} summarize performance
across all fine-tuned variants.
Fine-tuning with 100--150 examples yields consistent gains in round-trip fidelity
without significant runtime or token overhead.
Beyond this scale, returns diminish slightly, confirming that compact domain
adaptation suffices for high logical precision.

\begin{table}[t]
\centering
\caption{Summary of fine-tuning performance on the 201-spec subset. 
All values are averaged per item.}
\label{tab:ft-summary}
\resizebox{\columnwidth}{!}{%
\begin{tabular}{lcccc}
\toprule
\textbf{Model} & \textbf{Mean Sim. (\%)} & \textbf{Median Sim. (\%)} &
\textbf{Mean Runtime (s)} & \textbf{SAT/UNSAT Correctness (\%)}\\
\midrule
Baseline & 80.2 & 79.4 & 25.81 & 91.8\\
FT-50 & 85.1 & 86.3 & 37.96 & 93.9\\
FT-100 & 87.2 & 88.1 & 28.24 & 94.5\\
FT-150 & \textbf{87.7} & \textbf{88.4} & 23.38 & \textbf{94.5}\\
FT-300 & 87.6 & 88.2 & 19.27 & 94.3\\
\bottomrule
\end{tabular}}
\end{table}

\begin{figure}[t]
\centering
\safeincludegraphics[width=\columnwidth]{ft_scaling_curve_new.png}
\caption{Fine-tuning efficiency--fidelity trade-off. 
Compact fine-tuning (100--150 examples) improves fidelity 
with negligible latency increase.}
\label{fig:ft_scaling_curve}
\vspace{-5pt}
\end{figure}

These results confirm that small-scale domain adaptation effectively regularizes the model,
yielding more consistent transformations while preserving deterministic execution.

\vspace{3pt}
\subsection{Runtime Analysis}

\begin{figure}[t]
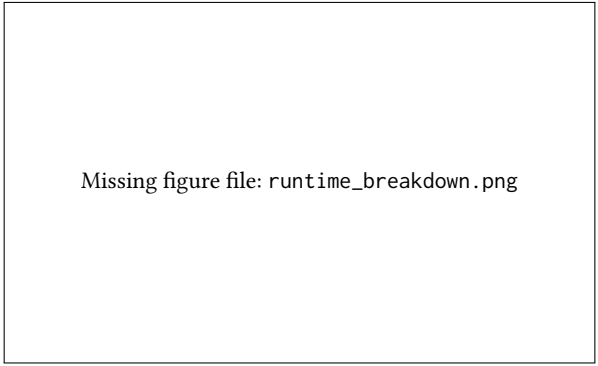

\label{fig:runtime_comparison}
\centering
\safeincludegraphics[width=\columnwidth]{runtime_breakdown.png}
\caption{Runtime breakdown by translation direction.
Validator overhead stays below 15\% of total pipeline runtime.}
\label{fig:runtime_breakdown}
\vspace{-5pt}
\end{figure}

Figure~\ref{fig:runtime_breakdown} presents the runtime breakdown by translation direction.
The NL$\rightarrow$PL step dominates at smaller datasets, while PL$\rightarrow$NL converges
faster as fine-tuning increases. 
Validator overhead remains below 15\% of total execution time,
demonstrating that semantic verification is efficient relative to translation cost.

\begin{figure}[h]
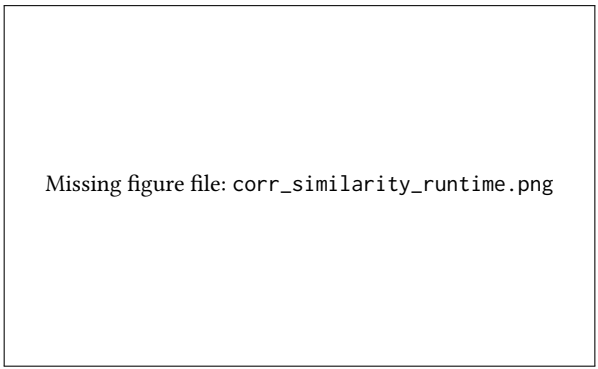

 \centering
 \safeincludegraphics[width=\columnwidth]{corr_similarity_runtime.png}
 \caption{Correlation between runtime and round-trip similarity
 across fine-tuned models (\(|r|<0.3\)).}
 \label{fig:corr-sim-runtime}
\end{figure}

End-to-end execution across all fine-tuned models is summarized in
Fig.~\ref{fig:corr-sim-runtime}. The near-flat trend
($|r|{<}0.3$ correlation between runtime and fidelity)
shows that higher precision does not increase latency.


\vspace{3pt}

\subsection{Reliability--Coverage Frontier}

\begin{figure}[t]
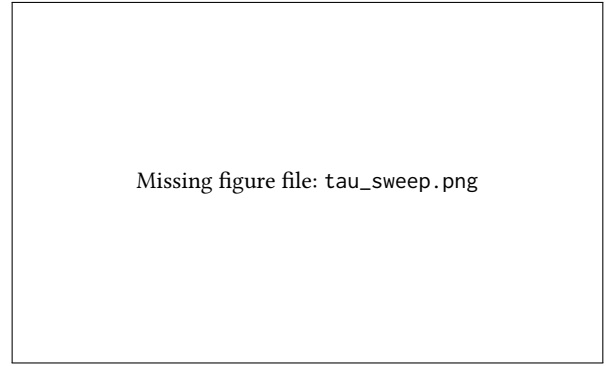

\centering
\safeincludegraphics[width=\columnwidth]{tau_sweep.png}
\caption{Reliability--coverage frontier across validator thresholds $\tau$; at $\tau = 75$, we retain 67.7\% of instances with 94.2\% SAT/UNSAT accuracy.}
\label{fig:tau_sweep_frontier}

\vspace{-5pt}
\end{figure}

As shown in Figure~\ref{fig:tau_sweep_frontier} and Table~\ref{tab:appendix-equiv}, 
high-confidence thresholds ($\tau\!\ge\!80$) maintain over 95\% CNF equivalence, 
while moderate ranges ($70\!\le\!\tau\!<\!80$) still exceed 85\% accuracy, 
demonstrating that validator-gated acceptance effectively isolates reliable translations.

\begin{table}[h]
\centering
\caption{CNF equivalence breakdown by similarity threshold bin.}
\label{tab:appendix-equiv}
\small
\begin{tabular}{lccc}
\toprule
 $\tau$ Range & Support & Equiv. Rate & Notes \\
 \midrule
 $\ge 90$ & 800 & 98.3\% & Ultra-high fidelity \\
$[$80, 90) & 471 & 95.6\% & High confidence \\
$[$70, 80) & 277 & 85.2\% & Moderate \\
 $<70$ & 552 & 54.1\% & Low fidelity \\
 \bottomrule
\end{tabular}
\vspace{-4pt}
\end{table}
\vspace{3pt}

\noindent\textbf{Validator threshold sweep.}
Coverage is the fraction of instances retained after filtering by NL$\leftrightarrow$PL consistency (similarity $\ge \tau$). Correctness reports SAT/UNSAT accuracy on the retained subset. We sweep the validator threshold $\tau$ from 60 to 95 to study the reliability--coverage trade-off. Increasing $\tau$ makes the validator more selective, so coverage monotonically decreases as fewer instances pass the NL$\leftrightarrow$PL consistency filter. At the same time, the SAT/UNSAT accuracy on the retained subset remains stable around $\sim$94\%, indicating that the validator primarily controls how much data we keep rather than substantially changing the correctness of accepted predictions; we use $\tau = 75$ as a balanced operating point (67.7\% coverage, 94.2\% SAT/UNSAT accuracy).

\subsection{Qualitative Solver Outcomes}

To contextualize the quantitative metrics, we include in the supplementary materials (Figure 1 under ``A.2 Illustrative End-to-End Verification Examples'') two illustrative examples covering both solver states (SAT and UNSAT), demonstrating how VeriTrans preserves semantic alignment while maintaining solver-verified determinism.

Beyond aggregate metrics, we verified that VeriTrans's fine-tuning 
improvements are statistically reliable rather than artifacts of sample variance. 
Following MLSys reproducibility practices, we applied two complementary tests: 
the non-parametric Wilcoxon signed-rank test to assess paired median shifts 
and a bias-corrected accelerated (BCa) bootstrap to estimate confidence intervals. 
As summarized in \autoref{sec:stats}, both analyses confirm that 
fine-tuned variants yield significant and consistent fidelity gains 
($p<10^{-5}$) over the baseline, with medium effect sizes.

\subsection{Statistical Validation}
\label{sec:stats}
Fine-tuned models significantly outperform the baseline. 
Tables~\ref{tab:wilcoxon}--\ref{tab:bootstrap} report the Wilcoxon signed-rank and bootstrap tests, both showing $p<10^{-5}$.

\begin{table}[h]
\centering
\caption{Wilcoxon signed-rank test comparing fine-tuned models with baseline.}
\label{tab:wilcoxon}
\small
\begin{tabular}{lccc}
\toprule
 Model & Median $\Delta$ (\%) & $p$-value & Cliff's $\delta$ \\
 \midrule
 FT-50 & +0.9 & $<10^{-5}$ & 0.35 (small) \\
 FT-100 & +1.4 & $<10^{-5}$ & 0.42 (medium) \\
 FT-150 & +1.6 & $<10^{-5}$ & 0.44 (medium) \\
\bottomrule
\end{tabular}
\vspace{-4pt}
\end{table}

 \begin{table}[h]
 \centering
 \caption{BCa bootstrap (10 000 resamples) 95\% confidence intervals for mean similarity gain.}
 \label{tab:bootstrap}
 \small
 \begin{tabular}{lcc}
 \toprule
 Model & Mean $\Delta$ (\%) & 95\% CI (BCa) \\
 \midrule
 FT-50 & +0.9 & $[$ +0.6 , +1.2 $]$ \\
 FT-100 & +1.4 & $[$ +1.1 , +1.7 $]$ \\
 FT-150 & +1.6 & $[$ +1.2 , +1.9 $]$ \\
 \bottomrule
\end{tabular}
\vspace{-4pt}
\end{table}

These findings verify that VeriTrans's improvements are not random fluctuations 
but systematic gains arising from domain-specific fine-tuning. 
The convergence of results from independent tests ($p<10^{-5}$) demonstrates 
the robustness of the approach and supports its reproducibility claims. 
We next discuss how these verified reliability properties extend to broader 
deployment scenarios and system-level performance.

\section{Related Work}
\label{sec:relatedwork}

Our work builds upon four converging research threads: 
(1) natural-language to formal-logic translation, 
(2) LLM--symbolic reasoning integration, 
(3) LLMs for formal verification and specification, and 
(4) hybrid neuro-symbolic reasoning systems for reliable automation.

\subsection{Natural Language to Logic Translation}

Recent studies increasingly focus on converting natural-language requirements into formal logic. 
Early systems relied on templates or semantic parsers, while newer approaches use instruction-tuned LLMs for flexible logical-form generation. 
\citet{yang2024harnessing} and \citet{han2024folio} demonstrate mapping from English to First-Order Logic (FOL) with strong generalization, and \citet{xu2024symboliccot} propose symbolic chain-of-thought prompting to improve reasoning faithfulness. 
Datasets such as FOLIO~\citep{han2024folio} and P-FOLIO~\citep{pfolo2024} enable fine-grained fidelity analysis, while NL2LTL~\citep{fuggitti2023nl2ltl}, SYNTHTL~\citep{mendoza2024synthtl}, and VLTL-Bench~\citep{english2025vltlbench} target temporal logic. 
These efforts expand linguistic coverage but seldom evaluate throughput, latency, or deterministic replay -- core system metrics for reproducible ML pipelines. 
\textbf{VeriTrans} extends this line of work by implementing a complete NL$\!\rightarrow\!$PL$\!\rightarrow\!$CNF$\!\rightarrow\!$SAT pipeline that integrates round-trip validation and exposes tunable fidelity--latency trade-offs.

\subsection{LLMs with Symbolic Solvers and Formal Tools}

Another active direction couples LLMs with symbolic reasoning engines. 
Logic-LM~\citep{pan2023logiclm}, SatLM~\citep{ye2023satlm}, and Symbolic-CoT~\citep{xu2024symboliccot} interleave text generation with solver feedback to maintain local consistency. 
\citet{ryu2025divide} and \citet{qi2025llmSymbolic} explore solver-augmented pipelines for formula validation, and \citet{qin2025llmverify,lam2024toolbased} examine such coupling in practical verification tasks. 
Unlike these solver-in-the-loop methods, \textbf{VeriTrans} externalizes solver feedback as a deterministic post-processing stage, enabling explicit profiling of validator latency and reliability under bounded compute. 
This design treats solver interaction as a measurable, reproducible systems component rather than an implicit prompt heuristic, improving both interpretability and reproducibility.

\subsection{LLMs for Formal Verification and Specification}

LLMs have also been applied to formal specification, program reasoning, and proof assistance. 
Most prior systems emphasize symbolic correctness but rarely disclose system-level metrics such as token efficiency or solver throughput. 
Work on requirements formalization~\citep{ferrari2025requirementsLLM,beg2025leveragingLLMFormalSpec} and logical synthesis~\citep{vossel2025advancing,sheng2025solsearch} demonstrates feasibility but not scalability. 
Similarly, frameworks integrating theorem provers~\citep{yang2023leandojo,wu2022autoformalization,lin2024fvel,hazra2025satquest} emphasize proof search over lightweight translation. 
\textbf{VeriTrans} complements these approaches by prioritizing reproducibility and runtime profiling, serving as a reliability-first front-end to solvers rather than a new proof engine.

\subsection{Neuro-Symbolic and Hybrid Reasoning}
Hybrid reasoning systems combine neural adaptability with symbolic guarantees. 
Adaptive symbolic selection~\citep{wang2025adaptiveSymbolic}, Mixture-of-Thought reasoning~\citep{zheng2025learningreasonmixtureofthoughtlogical}, and uncertainty-aware translation~\citep{wang2025conformalnl2ltl} show that symbolic priors reduce hallucination and improve trustworthiness. 
\citet{ryu2025divide} and \citet{hao2025planningSMT} further connect formal verification to high-level planning. 
\textbf{VeriTrans} extends this paradigm into a production-ready ML system: a compact fine-tuned model achieving high SAT/UNSAT correctness with low latency and fully reproducible execution.

\begin{table}[t]
\centering
\caption{At-a-glance contrast with representative prior systems.}
\label{tab:rw-contrast}
\vspace{3pt}
\setlength{\tabcolsep}{3.5pt}
\renewcommand{\arraystretch}{1.05}
\begin{adjustbox}{max width=\columnwidth}
\begin{tabular}{@{}lp{3.5cm}cclp{1cm}@{}}
\toprule
\textbf{System} & \textbf{Scope} & \textbf{Solver FB} & \textbf{Round-trip} & \textbf{Metrics} \\
\midrule
FOLIO~\citep{han2024folio} & NL$\rightarrow$FOL & --- & --- & --- \\
P-FOLIO~\citep{pfolo2024} & NL$\rightarrow$FOL & --- & --- & --- \\
Symbolic-CoT~\citep{xu2024symboliccot} & NL$\rightarrow$Logic (+solver) & Partial & --- & Limited \\
SatLM~\citep{ye2023satlm} & NL$\rightarrow$Logic (+solver) & Partial & --- & Limited \\
\textbf{VeriTrans} & \textbf{NL$\rightarrow$PL$\rightarrow$CNF$\rightarrow$SAT} 
& \textbf{Post-process} & \textbf{Yes} & \textbf{Fidelity/latency/tokens} \\
\bottomrule
\end{tabular}
\end{adjustbox}
\vspace{-6pt}
\end{table}

\paragraph{Contrast to prior systems.}
Unlike solver-in-the-loop prompting (e.g., Symbolic-CoT or SatLM), \textbf{VeriTrans} externalizes solver interaction as a deterministic post-processing stage and \emph{profiles it} as a first-class systems component (latency, coverage, tokens). 
Prior NL$\!\rightarrow\!$Logic efforts emphasize linguistic coverage but seldom report token efficiency, end-to-end throughput, or byte-identical replay. 
Our design operationalizes round-trip validation as a tunable acceptance policy (Sec.~\ref{sec:stage4-validators}) and reports how $\tau$ shifts correctness and coverage (Fig.~\ref{fig:tau_sweep_frontier}).

\section{Conclusion and Future Work}
\label{sec:conclusion}

This paper introduced \textbf{VeriTrans}, a lightweight neural--symbolic compiler
that translates natural-language requirements into verified propositional logic
representations. By coupling a fine-tuned LLM with deterministic CNF conversion
and solver-based validation, VeriTrans achieves over 94\% SAT/UNSAT correctness
and 87\% median round-trip fidelity on SatBench, confirming that small-scale
domain adaptation can deliver reliable logical translation without heavy
infrastructure. Fine-tuning with only 100--150 examples yields stable gains in
semantic precision while maintaining constant runtime and token efficiency,
supporting the claim that compact, domain-specialized models can rival large,
general-purpose systems when paired with formal verification feedback.

Beyond this work, we envision extending VeriTrans in several directions.
First, the translation and validation framework can naturally generalize from
propositional to first-order and temporal logics, enabling richer reasoning over
structured software specifications. Second, integrating reinforcement or
counterfactual feedback loops would allow the validator to guide model
correction dynamically, closing the gap between symbolic reasoning and learned
optimization. Finally, applying the same reliability-first paradigm to verified
compiler optimization or security-critical domains could establish a broader
class of LLM-assisted verification pipelines. Together, these directions
advance the long-term goal of building \emph{trustworthy machine learning
systems that reason, translate, and verify with a deterministic symbolic back-end and reproducible artifacts}.

\paragraph{Reproducibility and release.}
We will release prompts, deterministic seed/config files, CNF encoder, validators, and frozen per-item logs (tokens, time, hashes) together with an open-model replication suite and a replay script that regenerates byte-identical DIMACS outputs from a given PL formula, plus per-item logs to support end-to-end regression testing. Validator ablations and the $\tau$-sweep are included for direct reuse in reliability-critical NL$\rightarrow$logic front-ends.

\section*{Acknowledgments}
The authors used OpenAI’s ChatGPT to assist with rephrasing and restructuring text for clarity. All research contributions, dataset construction, experimental design, and analysis were carried out solely by the authors.

\section*{Data Availability}

The fine-tuning code and the complete set of experiments reported in this paper are available on Figshare at:
\url{https://figshare.com/s/2e7b48420f2517128c14}.

\IfFileExists{Reference_READY2.bib}{%
\bibliographystyle{ACM-Reference-Format}
\bibliography{Reference_READY2}

@article{hoeffding1963probability,
  title   = {Probability Inequalities for Sums of Bounded Random Variables},
  author  = {Hoeffding, Wassily},
  journal = {Journal of the American Statistical Association},
  volume  = {58},
  number  = {301},
  pages   = {13--30},
  year    = {1963}
}

@article{wei2022cot,
  title   = {Chain-of-Thought Prompting Elicits Reasoning in Large Language Models},
  author  = {Wei, Jason and Wang, Xuezhi and Schuurmans, Dale and Bosma, Maarten and Ichter, Brian and Xia, Fei and Chi, Ed and Le, Quoc and Zhou, Denny},
  journal = {arXiv preprint arXiv:2201.11903},
  year    = {2022}
}

@article{huang2025survey,
  title   = {A Survey on Hallucination in Large Language Models},
  author  = {Huang, Lei and Yu, Weijiang and Ma, Weitao and Zhong, Weihong and Feng, Zhangyin and Wang, Haotian and Chen, Qianglong and Peng, Weihua and Feng, Xiaocheng and Qin, Bing},
  journal = {ACM Transactions on Information Systems},
  volume  = {43},
  number  = {2},
  pages   = {1--55},
  year    = {2025}
}

@inproceedings{yang2024harnessing,
  title     = {Harnessing the Power of Large Language Models for Natural Language to First-Order Logic Translation},
  author    = {Yang, Yuan and Xiong, Siheng and Payani, Ali and Shareghi, Ehsan and Fekri, Faramarz},
  booktitle = {Proceedings of ACL 2024},
  year      = {2024}
}

@inproceedings{xu2024symboliccot,
  title     = {Faithful Logical Reasoning via Symbolic Chain-of-Thought},
  author    = {Xu, Jundong and Fei, Hao and Pan, Liangming and Liu, Qian and Lee, Mong-Li and Hsu, Wynne},
  booktitle = {Proceedings of ACL 2024},
  year      = {2024}
}

@inproceedings{han2024folio,
  title     = {FOLIO: Natural Language Reasoning with First-Order Logic},
  author    = {Han, Shiyu and Liu, Yixin and Zhang, Tianyu and Wang, Yuxin and Ren, Xiang and He, He},
  booktitle = {Proceedings of EMNLP 2024},
  year      = {2024}
}

@inproceedings{ryu2025divide,
  title     = {Divide and Translate: Compositional First-Order Logic Translation and Verification},
  author    = {Ryu, Hyun and Kim, Gyeongman and Lee, Hyemin S. and Yang, Eunho},
  booktitle = {ICLR 2025},
  year      = {2025}
}

@article{wei2025satbench,
  title   = {SATBench: Benchmarking LLMs' Logical Reasoning via Automated Puzzle Generation from SAT Formulas},
  author  = {Wei, Anjiang and Wu, Yuheng and Wan, Yingjia and Suresh, Tarun and Tan, Huanmi and Zhou, Zhanke and Koyejo, Sanmi and Wang, Ke and Aiken, Alex},
  journal = {arXiv preprint arXiv:2505.14615},
  year    = {2025}
}

@inproceedings{ye2023satlm,
  title     = {SatLM: Satisfiability-Aided Language Models Using Declarative Prompting},
  author    = {Ye, Xi and Chen, Qiaochu and Dillig, Isil and Durrett, Greg},
  booktitle = {NeurIPS 2023},
  year      = {2023}
}

@article{pan2023logiclm,
  title   = {Logic-LM: Empowering Large Language Models with Symbolic Solvers},
  author  = {Pan, Liangming and Albalak, Alon and Wang, Xiaoman and Wang, William Yang},
  journal = {arXiv preprint arXiv:2305.12295},
  year    = {2023}
}

@inproceedings{wu2022autoformalization,
  title     = {Autoformalization with Large Language Models},
  author    = {Wu, Yuhuai and Jiang, Albert Q. and Li, Wenda and Rabe, Markus N. and Staats, Charles and Jamnik, Mateja and Szegedy, Christian},
  booktitle = {NeurIPS 2022},
  year      = {2022}
}

@inproceedings{yang2023leandojo,
  title     = {LeanDojo: Theorem Proving with Retrieval-Augmented Language Models},
  author    = {Yang, Kexun and others},
  booktitle = {NeurIPS 2023},
  year      = {2023}
}

@inproceedings{lin2024fvel,
  title     = {FVEL: Interactive Formal Verification Environment with Large Language Models},
  author    = {Lin, Xuefeng and others},
  booktitle = {NeurIPS 2024},
  year      = {2024}
}

@inproceedings{qin2025llmverify,
  title     = {Can Large Language Models Verify System Software?},
  author    = {Qin, Jianxing and Du, Alexander and Zhang, Danfeng and Lentz, Matthew and Zhuo, Dapeng},
  booktitle = {HotOS 2025},
  year      = {2025}
}

@misc{openai_finetuning,
  author       = {{OpenAI}},
  title        = {OpenAI Fine-Tuning API Documentation},
  howpublished = {\url{https://platform.openai.com/docs/guides/fine-tuning}},
  year         = {2025}
}

@article{DUAN2025115368,
title = {SAT-based bounded model checking for propositional projection temporal logic},
journal = {Theoretical Computer Science},
volume = {1049},
pages = {115368},
year = {2025},
issn = {0304-3975},
author = {Zhenhua Duan and Cong Tian and Nan Zhang and Chaofeng Yu and Mengfei Yang and Jia He},


}

@article{beg2025leveragingLLMFormalSpec,
  title        = {Leveraging LLMs for Formal Software Requirements},
  author       = {Beg, A. and others},
  journal      = {arXiv preprint arXiv:2507.14330},
  year         = {2025},
  url          = {https://arxiv.org/abs/2507.14330}
}

@article{english2025vltlbench,
  title        = {Verifiable Natural Language to Linear Temporal Logic Translation: A Benchmark Dataset and Evaluation Suite (VLTL-Bench)},
  author       = {English, William H. and Walker, Chase and Simon, Dominic and Jha, Sumit K. and Ewetz, Rickard},
  journal      = {arXiv preprint arXiv:2507.00877},
  year         = {2025},
  url          = {https://arxiv.org/abs/2507.00877}
}

@article{ferrari2025requirementsLLM,
  title        = {Formal Requirements Engineering and Large Language Models: Opportunities and Challenges},
  author       = {Ferrari, Alessio and others},
  journal      = {Information and Software Technology},
  year         = {2025},
  doi          = {10.1016/j.infsof.2025.107697}
}

@inproceedings{fuggitti2023nl2ltl,
  title        = {NL2LTL: A Python Package for Converting Natural Language Instructions to Linear Temporal Logic},
  author       = {Fuggitti, Francesco and others},
  booktitle    = {AAAI 2023 (Demo Track)},
  year         = {2023},
  doi          = {10.1609/aaai.v37i13.27068}
}

@inproceedings{hao2025planningSMT,
  title        = {Large Language Models Can Solve Real-World Planning Rigorously with Formal Verification Tools},
  author       = {Hao, Yuwei and others},
  booktitle    = {NAACL 2025},
  year         = {2025}
}

@article{hazra2025satquest,
  title        = {SATQuest: A Verifier for Logical Reasoning Evaluation and Benchmarks},
  author       = {Hazra, S. and others},
  journal      = {arXiv preprint arXiv:2509.00930},
  year         = {2025},
  url          = {https://arxiv.org/abs/2509.00930}
}

@inproceedings{lam2024toolbased,
  title        = {A Closer Look at Tool-based Logical Reasoning with LLMs},
  author       = {Lam, Lok Hin Marcus and others},
  booktitle    = {ALTA 2024},
  year         = {2024},
  url          = {https://aclanthology.org/2024.alta-1.4.pdf}
}

@inproceedings{mendoza2024synthtl,
  title        = {Translating Natural Language to Temporal Logics with Large Language Models},
  author       = {Mendoza, Ricardo and Trippel, Caroline and others},
  booktitle    = {Formal Methods in Computer-Aided Design (FMCAD)},
  year         = {2024}}

@article{pfolo2024,
  title        = {P-FOLIO: Evaluating and Improving Logical Reasoning with First-Order Logic Proofs},
  author       = {Liu, Yixin and Han, Shiyu and Zhang, Tianyu and Ren, Xiang and He, He},
  journal      = {arXiv preprint arXiv:2410.09207},
  year         = {2024},
 url          = {https://arxiv.org/abs/2410.09207}
}

@inproceedings{qi2025llmSymbolic,
  title        = {Large Language Models Meet Symbolic Solvers for Faithful Logical Reasoning},
  author       = {Qi, Chen and Zhang, Xinyu and Guo, Jiayi and Han, Dongxu},
  booktitle    = {ICLR 2025},
  year         = {2025}
}

@inproceedings{sheng2025solsearch,
  title        = {An LLM-Driven Framework for Efficient SAT-Solving Code Search and Optimization},
  author       = {Sheng, J. and others},
  booktitle    = {ICSE NIER 2025},
  year         = {2025},
  doi          = {10.1109/ICSE-NIER66352.2025.00007},
  url          = {https://dl.acm.org/doi/abs/10.1109/ICSE-NIER66352.2025.00007}
}

@article{vossel2025advancing,
  title        = {Advancing Natural Language Formalization to First Order Logic with Fine-tuned LLMs},
  author       = {Vossel, Felix and Mossakowski, Till and Gehrke, Bj\"orn},
  journal      = {arXiv preprint arXiv:2509.22338},
  year         = {2025},
  url          = {https://arxiv.org/abs/2509.22338}
}

@article{wang2025adaptiveSymbolic,
  title        = {Adaptive Selection of Symbolic Languages for Improving LLM Logical Reasoning},
  author       = {Wang, Xiangyu and Yang, Haocheng and Cheng, Fengxiang and Liu, Fenrong},
  journal      = {arXiv preprint arXiv:2510.10703},
  year         = {2025},
  url          = {https://arxiv.org/abs/2510.10703}
}

@article{wang2025conformalnl2ltl,
  title        = {ConformalNL2LTL: Uncertainty-Aware Natural Language to Temporal Logic Translation with Large Language Models},
  author       = {Wang, J. and others},
  journal      = {arXiv preprint arXiv:2504.21022},
  year         = {2025},
  url          = {https://arxiv.org/abs/2504.21022}
}

@misc{zheng2025learningreasonmixtureofthoughtlogical,
      title={Learning to Reason via Mixture-of-Thought for Logical Reasoning}, 
      author={Tong Zheng and Lichang Chen and Simeng Han and R. Thomas McCoy and Heng Huang},
      year={2025},
      eprint={2505.15817},
      archivePrefix={arXiv},
      primaryClass={cs.CL},
      url={https://arxiv.org/abs/2505.15817}, 
}
}{%

}

\end{document}